\documentclass[preprint,12pt,authoryear]{elsarticle}

\usepackage{amssymb}
\usepackage{amsmath}
\usepackage{graphicx}
\usepackage{multirow}
\usepackage{amsmath,amssymb,amsfonts}
\usepackage{amsthm}
\usepackage{mathrsfs}
\usepackage[title]{appendix}
\usepackage{xcolor}
\usepackage{colortbl}
\usepackage{textcomp}
\usepackage{manyfoot}
\usepackage{booktabs}
\usepackage{algorithm}
\usepackage{algorithmicx}
\usepackage{algpseudocode}
\usepackage{listings}
\definecolor{mistyrose}{rgb}{1.0, 0.89, 0.88}

\newcommand{\net}{ChemNNE\ }

\journal{Neural Networks}

\begin{document}

\begin{frontmatter}



\title{Neural Network Emulator for Atmospheric Chemical ODE} 

\author[a,b]{Zhi-Song Liu}
\author[c]{Petri Clusius}
\author[b,c]{Michael Boy}

\affiliation[a]{organization={School of Engineering Science, Lappeenranta-Lahti University of Technology LUT},
            city={Lahti},
            postcode={15110},
            country={Finland}}
            
\affiliation[b]{organization={Atmospheric Modelling Centre Lahti, Lahti University Campus},
            city={Lahti},
            postcode={15140},
            country={Finland}}
            
\affiliation[c]{organization={Institute for Atmospheric and Earth System Research (INAR), The University of Helsinki},
            city={Helsinki},
            postcode={00014},
            country={Finland}}



\begin{abstract}
Modelling atmospheric chemistry is complex and computationally intense. Given the recent success of Deep neural networks in digital signal processing, we propose a Neural Network Emulator for fast chemical concentration modelling. We consider atmospheric chemistry as a time-dependent Ordinary Differential Equation. To extract the hidden correlations between initial states and future time evolution, we propose \net, an Attention based Neural Network Emulator (NNE) that can model the atmospheric chemistry as a neural ODE process. To efficiently capture temporal patterns in chemical concentration changes, we implement sinusoidal time embedding to represent periodic tendencies over time. Additionally, we leverage the Fourier neural operator to model the ODE process, enhancing computational efficiency and facilitating the learning of complex dynamical behavior. We introduce three physics-informed loss functions, targeting conservation laws and reaction rate constraints, to guide the training optimization process. To evaluate our model, we introduce a unique, large-scale chemical dataset designed for neural network training and validation, which can serve as a benchmark for future studies. The extensive experiments show that our approach achieves state-of-the-art performance in modelling accuracy and computational speed.
\end{abstract}


\end{frontmatter}



\section{Introduction}\label{introduction}

In the past decades, deep learning has been proven as an efficient data-driven modelling approach that can handle large-scale datasets for complex digital signal processing, like images (\citet{dalle,styledalle}), videos (\citet{adversarial}), audio (\citet{audio}), and languages (\citet{gpt2,gpt4,llama,llama2}). It is intriguing to see if it can benefit chemical modelling problems. There are a number of approaches that use deep neural networks to resolve specific chemical problems, like \citet{chemistry1,chemistry2}. In general, we can take the time evolution of the chemical compound concentration as an Ordinary Differential Equation (ODE), which can be modeled by neural networks. The underlying dynamics of chemical reactions are implicitly captured and optimized through training on data. In this way, we can build neural networks as surrogate models to replace numerical simulations for faster computation. The development of fast neural emulators also helps address the computational bottlenecks in large-scale, high-resolution chemical simulations. Specifically, within global climate models containing millions of grid cells, a fast neural network can accelerate the modeling process at the grid-cell level. This approach significantly reduces computation time by aggregating information from all grid cells to estimate crucial chemical processes relevant to high-resolution climate predictions, enabling near real-time, high-resolution forecasting. While similar research has been conducted in~\citet{chem1,chem2}, few studies address large-scale atmospheric chemistry modelling or tackle multiple-input-and-multiple-output chemical predictions. Therefore, in this study, we aim at an end-to-end efficient neural network that can model multiple chemical concentrations (49$\sim$300 aerosol chemical compounds) over time (one hour). Specifically, given environmental parameters and initial chemical concentrations, the model can predict both the future states of these compounds and the emergence of new compounds from chemical interactions. We evaluate model performance by analyzing temporal concentration changes, using mean squared errors for accuracy and running time for computational efficiency. To interpret the learning ability of attention, we also analyze the key chemical components for their effects and correlations to other compounds. To the best of our knowledge, our proposed \net is the first neural network-based approach for atmospheric chemical emulation, addressing large-scale multi-compound predictions over time. To summarize, our contributions are:

\begin{itemize}
\item We propose the first neural network emulator for atmospheric chemical modelling, designed to predict the future evolution of chemical concentrations based on initial chemical concentrations and environmental parameters.
\item To model the inter- and intra-correlations among chemical data, we propose to use time-embedded attention to model the molecule concentration as a learnable time-dependent process.
\item We combine the Fourier Neural Operator (FNO) and attention to represent chemical signals in both spatial and frequency domains, facilitating a computationally efficient neural ODE model.
\item To better constrain the chemical process, we propose physics-informed losses, including identity loss, derivative loss and mass conservation loss to ensure that the model can predict further future steps without violating the chemistry.
\item We conduct extensive experiments on three large-scale tasks, including interperiod, intraperiod and hybridperiod tasks, to show the efficiency of our \net on various atmospheric chemical problems.
\end{itemize}

\section{Related Work}\label{related_work}
In atmospheric chemistry, a key focus is studying the oxidation of volatile organic compounds to understand atmospheric transformations. With prior knowledge of the emitted chemical compounds and the main atmospheric reaction paths, researchers analyze atmospheric phenomena from both a micro view, focusing on the concentrations of trace volatile organic compounds (in the ppt or ppb range), and a macro view, examining large-scale effects like cloud formation and air pollution driven by particle formation and growth. Together, these perspectives provide a comprehensive understanding of atmospheric processes. From the mathematical perspective, we study the problem as a time-series problem or partial differential equation (PDE) problem. Given the booming development of deep learning, we will discuss deep learning based chemical modelling in the following sections.

\subsection{Deep learning for chemical modelling}
The rise of deep learning in the early 2010s has significantly expanded the scope of scientific discovery processes. Advances in GPU technology and innovative algorithms have enabled deep learning and artificial intelligence to integrate scientific knowledge effectively, supporting diverse tasks across multiple domains. AI for chemistry, one of the promising research directions, has recently attracted a lot of researchers for investigation, like weather forecasting, air quality estimation, aerosol chemical interaction, and so on. Graphcast (\citet{graphcast}) is one of the promising weather forecasting AI models that uses a graph neural network to predict the global weather with medium-range resolution. Recently, Aurora~\cite{aurora} has revolutionized many facets of climate studies by leveraging vast amounts of data and proposing a foundation model with 1.3B parameters, achieving state-of-the-art performance in weather forecasting. Unlike weather prediction models, atmospheric chemical models examine complex chemical dynamics, capturing interactions and transformations among compounds under physical constraints. It can be very useful for extreme weather study, air pollution prevention and so on. Bassetti et al.~\citet{diffesm} propose a conditional emulator via the conditional diffusion model to understand the impact of human actions on the earth system. In parallel, AI’s role in autonomous chemical design and optimization has gained attention. ChemGPT~(\cite{chemgpt}), fine-tuned from large language models like GPT-4~(\cite{gpt4}), facilitates reaction optimization and experimental automation. Air pollution mitigation, particularly through aerosol chemistry, is another focal area. For example, \cite{air_1} employs random forests to analyze PM2.5 levels during haze episodes pre- and post-COVID-19 lockdown, while \cite{air_2} uses graph machine learning to impute missing data in tropospheric ozone measurements, bridging gaps in atmospheric pollutant time series. Additional research has applied machine learning to model other pollutants, such as OH~(\cite{air_3}) and NO2~(\cite{air_4}). Recently, there has been growing interest in using neural network emulators as fast surrogate models to replace computationally intensive numerical simulations in chemical modelling. Not only can they be built into chemical systems for fast simulation and analysis, but they can also be used in high-resolution global climate modelling. For instance, \cite{emulator_1} demonstrates the effectiveness of attention models in simulating Atmospheric Chemistry Box Models, achieving speeds over two orders of magnitude faster than traditional solvers. Similarly, \cite{emulator_2} fine-tunes transformer models to address cheminformatics tasks. ~\cite{emulator_3} utilizes autoencoders and LSTMs for global chemical model simulations, while \cite{emulator_4} introduces a 28-layer residual regression neural network for chemical transport modelling. Despite these advancements, most of these approaches are tailored toward spatiotemporal chemical modelling for short-term weather or air quality analysis. Comprehensive investigations into detailed atmospheric chemical reactions remain relatively unexplored and present a critical area for future research.

\subsection{Neural autoregression and ODE for time series learning}
Given the initial concentrations of chemical compounds and environmental parameters, the overall process of estimating future chemical changes can be framed as either an Ordinary Differential Equation (ODE) approach or as a general time series model, depending on the complexity of temporal dynamics involved. There are two primary approaches for solving time series problems: (1) autoregressive models and (2) ordinary differential equation (ODE)-based processes. Though only a few works are directly related to chemical ODE modelling, summarizing time series works can help establish a foundation for understanding how chemical reactions can be processed and modeled by deep learning.

For autoregression, RNN~(\cite{lstm}) and attention (\cite{attention}) are two major approaches for temporal modelling. RNN-based methods capture sequential temporal variations through recurrent structures, where state transitions are inherently time-dependent. In contrast, attention-based or transformer-based approaches capture global temporal dependencies, identifying specific time steps that are most relevant for accurate future prediction. Specifically, \cite{logform,informer} propose efficient self-attention models that achieve strong forecasting performance by focusing on scalable attention mechanisms. \cite{autoformer} introduces Autoformer, which leverages Fast Fourier Transform to compute auto-correlations within time series data, enhancing its ability to capture periodic patterns. FEDformer (\cite{fedformer}) further improves forecasting performance by incorporating a mixture-of-experts design, which ensembles specialized knowledge learned from the frequency domain to enhance the model’s ability to capture complex temporal patterns. In cases with partial time series data, TimeNet (\cite{timesnet}) proposes the TimesBlock, which projects 1D time series data into a 2D space to capture inter- and intra-time correlations, facilitating more nuanced temporal dependency modelling for improved forecasting.

Scientific hypotheses in fields like physics or chemistry are often expressed as discrete entities, such as symbolic formulas or chemical compounds. However, these can be represented in a differentiable space, making them suitable for efficient optimization using gradient-based methods. Different from general time series processing, neural ODE/PDE models offer explicit differential systems for optimization. The introduction of neural ODEs~(\cite{neuralode}) has unlocked new avenues for developing novel ODE solvers, with applications spanning physics~(\cite{physics1,physics2}), chemistry~(\cite{meshfreeflow,chemistry1,chemistry2}), and biology~(\cite{biology1,biology2}), among other fields. Deep learning-based ODE processes can be broadly categorized into three primary branches: (1) continuous learning~(\cite{physicsloss1,physicsloss2,physicsloss3,ode6}), (2) discrete learning~(\cite{discrete1,discrete2,discrete3,ode2,ode3,ode5}), and (3) neural operators~(\cite{fno,no,gno,fno}). In continuous learning, neural networks are used as domain projectors, mapping data into a continuous latent space to capture global solution patterns. This approach embeds physics-informed loss functions into the training process to ensure that the network’s behavior adheres to the underlying ODE dynamics. Discrete learning, on the other hand, models nonlinear dynamics on mesh grids by utilizing convolutional or graph neural networks.

\subsection{Neural operator for spatial and temporal encoding}
The recent progress in neural operators for differentiable signal representations~(\cite{siren,wire,bacon}) shows promise for predicting continuous solutions to ODE/PDE systems based on spatiotemporal queries. It is known that using MLP for 1D convolution causes bias in the low frequency domain~(\cite{bias}). Neural operators solve it by providing frequency-aware parameterization, such that high-frequency information will not be suppressed. Siren~(\cite{siren}) was one of the first methods to apply sinusoidal functions in neural networks for effective signal reconstruction, creating a foundation for continuous function representation. Fourier Neural Operators (FNO)~(\cite{fno}) advance this by using the Fast Fourier Transform to map signals in the frequency domain, parameterizing the integral kernel as a straightforward multiplication in Fourier space, significantly reducing computation. It parameterizes the integral kernel directly in Fourier space as simple multiplication, which is a significant computation reduction. For time series processing, a learnable time encoding system allows for flexible alignment of temporal features, enhancing the model’s adaptability to various time scales and improving temporal resolution. For instance, Spherical Fourier Neural Opeartor~(\cite{sfno}) generalizes the Fourier Transform as a spherical geometry, achieving stable autoregressive on long-term forecasting. \cite{gaia} and ~\cite{fourcastnet} are the two most recent works that use the Fourier Neural operator for global weather forecasting and chemical modelling. In this work, we aim to explore this technique further, combining it with existing ODE methods to achieve more robust and accurate modelling.

\section{Method}\label{method}

\subsection{Data preparation}
To understand real-world atmospheric chemistry, we need to generate a large-scale dataset for neural network training. Environmental chambers and flow reactors are used to study the complex chemical reactions in the atmosphere under well-controlled conditions. Such experiments have significantly improved our understanding of atmospheric chemical processes and the fate of many volatile organic compounds (VOCs), including anthropogenic pollutants. This has increased the theoretical understanding of atmospheric chemistry, leading to numerical chemical schemes which can describe the time evolution of the atmospheric chemistry.  To obtain the dataset, we use the ARCA-box model (\citet{arca}), a comprehensive toolbox for modelling atmospheric chemistry and aerosol processes. The model can be set up to any kinetic chemistry system, but most often the base chemistry scheme comes from the Master Chemical Mechanism (\cite{mcm,mcm1,mcm2}). We used a subset of the MCM, augmented with the Peroxy Radical Autoxidation Mechanism (\cite{pram}), which simulates monoterpene autoxidation, crucial in producing low volatility vapors from biogenic volatile organic carbon (BVOC) emissions. The subset was selected so that it contains both biogenically and anthropogenically emitted compounds such as isoprene, monoterpenes, aromatics, alkenes and alkynes, and together with their reaction products contains altogether 3301 compounds and 9530 reactions. The kinetic chemistry system is converted to a system of ODEs using the Kinetic PreProcessor (\cite{kpp}), and then incorporated in ARCA box and solved numerically to simulate the time series of the reaction products. 

To create the dataset, we first selected representative initial concentrations of 49 precursor gas compounds and three environmental conditions: temperature, relative humidity and short-wave radiation, affecting the reaction rates and photochemistry. Chemically related compounds were linked together and varied in tandem, resulting in 15 independently varied groups of initial concentrations. In other words, the variation of initial concentrations of 49 input chemical compounds is determined by 15 independent factors. These factors were selected so that they produced realistic lower and upper bounds of the initial concentrations, and thus obtained $2^{15}$ combinations of the initial concentrations. Additionally, all simulations were repeated with the environmental variables in the middle values. This produced $3^3$ different combinations of physical factors. Hence we obtain total $2^{15}\times 3^3=884736$ different chemical reaction samples. For each 1-hour time series, we sample the concentrations every 5 minutes, totalling 12 time steps. During the 1-hour simulation, new chemical compounds are generated, and including the original 49 input chemical compounds, we observe a total of 3301 chemical compounds. For our experiments, we focus on 400 of the most important chemical compounds and ignore others.

\begin{figure}[t]
	\centering
		\centerline{\includegraphics[width=\textwidth]{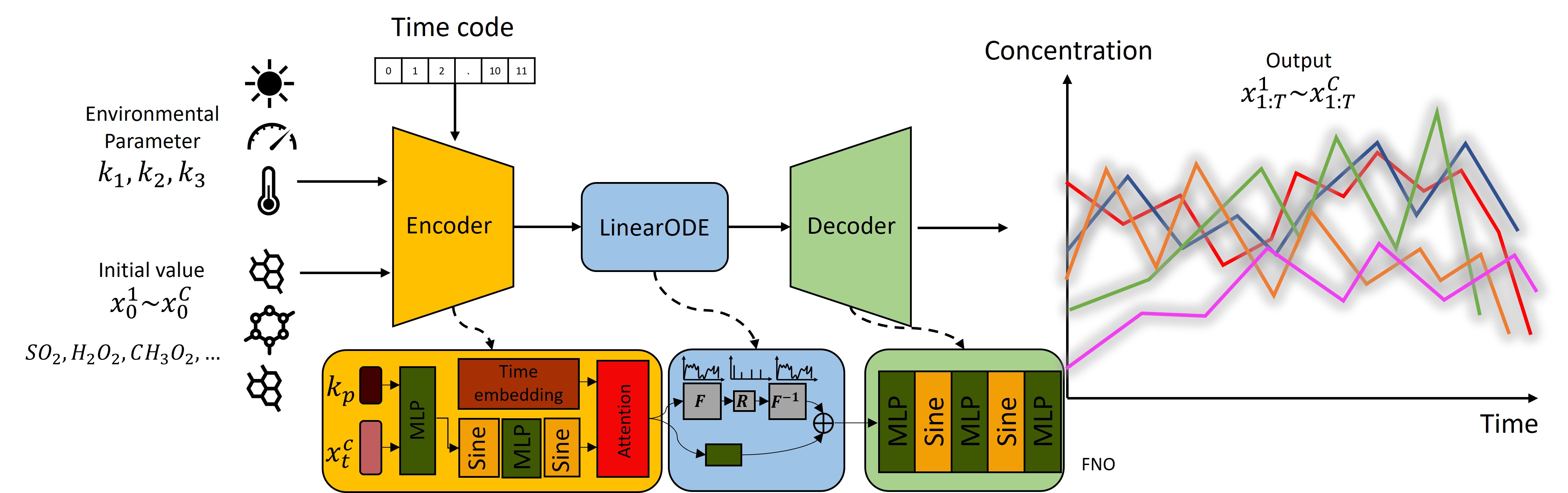}}
		\caption{\small{\textbf{The proposed \net for chemical concentration prediction.} It takes the environmental parameters and chemical initial concentration to predict the future chemical reaction process. 
		}}
		\label{fig:network}
\end{figure}

\subsection{\net for Atmospheric modelling}
To model the atmospheric chemical reaction, we propose the following framework in Figure~\ref{fig:network}. It consists of three parts: 1) encoder, 2) linearODE, and 3) decoder. Let us denote the initial chemical concentration as $x_0\in \mathbb{R}^N$, where $N$ is the number of chemical elements, and the environmental parameters as $k\in \mathbb{R}^O$, where $O$ is the number of environmental factors, e.g., temperature, humidity and radiation. Mathematically, given the time steps $t_i=0,1,...,T$, the proposed \net learns a linear ODE function that predicts time-dependent output chemical values $x_t\in \mathbb{R}^N$ as,

\begin{small}
\begin{equation}
x_t=\psi\left(\varphi(x_0,k)+\int_{[t_0,t_T]}f(z(t))dt\right),\ \text{where}\ f(z(t))=\frac{dz(t)}{dt}
\label{eq:ode}
\end{equation}
\end{small}

\noindent In Eq~\ref{eq:ode}, it comprises three components: an encoder $\varphi$, the linearODE $f$, and a decoder $\psi$. The dynamics of the chemical process are characterized as $z(t)$. Given initial chemical concentration $x_0$, we have the latent space initial state $z_0=\varphi(x_0,k)$, which serves as the initial condition for the ODE. The integral operation is evaluated by the proposed linearODE, which is achieved by the Fourier Neural Operator. 

To obtain continuous, differentiable signal representation, we apply sine as a periodic activation function to the chemical input and hidden features, $x_i \rightarrow \phi_i(x_i)=sin(W_i x_i + b), \text{where\ } W_i\sim\mathcal{U}(-6/\sqrt{N}, 6/\sqrt{N}), c\in \mathbb{R}$, as they are continuous changes in the physical world and it is memory efficient to compute the derivatives in the sine space without being constrained by discrete data samples. The initialized weight parameters ensure that the input to each sine activation is normally distributed with a standard deviation of 1. Physically, we can interpret the weights of the sinusoid function as angular frequencies while the biases are phase offsets. Applying the sine function keeps signal amplitude constant while expanding frequency bands for high-frequency modelling. The Encoder is made of sine-based MLP layers and time-dependent attention modules for extracting hidden correlations among different chemical components, and the decoder also uses the sine-based MLP layers for time prediction.

\subsubsection{Sinusoid function for time embedding} 
Observing the chemical reaction simulation, we can see that the chemical compounds usually dynamically change their concentrations over time. This oscillating behavior can resonate with radio frequency modulation, where the true signal can be encoded to the carrier wave by either changing its frequency or amplitude. As pointed out by \citet{siren}, using neural networks for continuous and differentiable physical signals is challenging because the neural networks tend to oversmooth the high-frequency details and do not represent the derivatives of a target signal well. To introduce a differentiable time representation, we propose to use sinusoidal time embedding (\citet{nerf}) to the time code, which can be described as follows.

\begin{small}
\begin{equation}
\lambda(t) = F_\Theta \circ t = \left( sin(2\Theta \pi t), cos(2\Theta \pi t) \right)
\label{eq:time}
\end{equation}
\end{small}

\noindent Mathematically, given the time steps $t_i=0,1,...,T$, we project them onto a higher dimensional space $\mathbb{R}^{2L}$, where $F_\Theta\in \mathbb{R}^L$ is the L-length learnable parameters that can define the frequency of the time code. We represent the time as a combination of sine and cosine operators so that the network can learn to adjust the reaction frequency.

\subsubsection{Attention for chemical representation} 
Attention has been widely used in image, video, and language processing. Its success comes from its efficient nonlocal feature representation. For chemical modelling, we also expect an efficient approach that can model the long-term time evolution, so that similar chemical behavior across time or different chemical compounds can be utilized for pattern matching. In order to achieve that, we propose a time-dependent attention module, which can encode and decode the chemical data to the latent space for implicit neural representation. Mathematically, we can define the process as,

\begin{small}
\begin{equation}
z_\lambda(t) = z_\lambda(t) + \text{ffn}\left(z_\lambda(t) + \sigma \left(\frac{Q_z K_z^T}{\sqrt{d}}\right) V_z \right)
\label{eq:attn}
\end{equation}
\end{small}

\noindent where $z_\lambda(t) = sin\left[W(z_0)+b\right]+\lambda(t)$, $Q_z=z_\lambda(t)W_Q$, $K_z=z_\lambda(t)W_K$, and $V_z=z_\lambda(t)W_V$. $\sigma$ is the softmax function. The $W_*$ is the learnable parameter. $z_0$ is the output feature of the Sinusoid mapping layer. $d$ is the dimension of the learned feature vector. As depicted in Figure~\ref{fig:network}, given the latent space chemical features, we apply the Sinusoid mapping first. Then we use the attention module to learn the nonlocal correlations. A feedforward network ($\text{ffn}$) is used to learn the residues for network update. Note that the activation of $\text{ffn}$ is also a sine function to preserve the periodic behavior of chemical reactions.

\subsubsection{Fourier Neural Operator (FNO)} 
To explore the derivatives of the chemical reactions, introducing ODE in the neural network can preserve the underlying physical laws. In recent works~\citet{neuralode,gno,fno}, neuralODE, like Torchdiffeq (\citet{neuralode}), has been used for GPU based ODE approximation. However, the disadvantage is that it is not numerically stable and rather slow in computation. We propose to further simplify the ODE process as a Neural Fourier Operation. Given the discrete-time representation across multiple chemical components, we use Fourier Transformation to project the signal to the frequency domain and learn the multi-grid nonlinear interactions across temporal and component domains.

\begin{small}
\begin{equation}
\begin{matrix}
\!\begin{aligned}
& z_{t+1}=\gamma \left( Wz_{t} + \mathcal{F}^{-1} (R_\phi \cdot (\mathcal{F}z_{t})) \right) \\
& (R_\phi \cdot (\mathcal{F}z_{t})) = \sum_{j=1}^{d_v} R_{k,l,j}{\mathcal{F}z_{t}}_{k,j}, k=1,...,k_{max}, j=1,...,d_v
\label{eq:fno}
\end{aligned}
\end{matrix} 
\end{equation}
\end{small}

\noindent We lift the signal to the frequency domain, given k frequency modes, we have $\mathcal{F}z_{t}(k)\in \mathbb{C}^{d_v}$ and $R_\phi(k)\in \mathbb{C}^{d_v \times d_v}$. $\gamma$ is the nonlinear activation. We define $R_\phi(k)$ as the truncation function that only keeps the maximal number of modes $k_{max}=|{k\in \mathbb{Z}_d: |k_j| \leq k_{max,j}, for j=1,...,d}|$.

The advantages of using FNO for ODE computation are 1) CNN and transformers can only learn on fixed-scale grids and fail to capture the fine scales of multi-scale systems, while FNO is grid-invariant since the operations outside of the Fourier layers act point-wise on the spatial domain, 2) it is faster to use Fourier Transform than convolution, as it is quasilinear, where the full standard integration of n points has complexity $O(n^2)$; 3) the input and outputs of PDEs are discrete functions, so it is efficient to represent them in the Frequency domain for global convolution. In the experiments, we would show the complexity comparisons to demonstrate the efficiency of FNO.

\subsubsection{Physics-informed losses.}
To train the whole \net, not only do we utilize the commonly used Mean Squared Errors (MSE) between prediction and ground truth, but we also propose to utilize the first- and second-order derivation, and total mass conservation loss. Mathematically, we can define the overall losses as

\begin{small}
\begin{equation}
Loss=\alpha_1 L_{recon}+\alpha_2 L_{d1}+\alpha_3 L_{d2}+\alpha_4 L_{idn}+\alpha_5 L_{mass}
\label{eq:loss}
\end{equation}
\end{small}

\noindent In Eq~\ref{eq:loss}, $\alpha_1$ to $\alpha_5$ are the weighting parameters that balance all five loss terms. $L_{recon}$ is the reconstruction loss measures the discrepancy between the prediction and ground truth as $L_{recon}=|x(t)-x'(t)|^2$. The first-order gradient loss enforces the predicted trajectories to closely follow the ground truth over time as $L_{d1}=|\frac{dx(t)}{dt}-\frac{dx'(t)}{dt}|^2$. Similarly, we can define the second-order gradient loss as $L_{d2}=|\frac{dx^2(t)}{dt^2}-\frac{dx'^2(t)}{dt^2}|^2$. Meanwhile, we can also enforce that the \net should preserve the initial condition unchanged during the training process. Hence we can define the identity loss as $L_{idn}=|x(t_0)-\psi(\varphi(x(t_0)))|^2$. Finally, we define the total mass conservation loss, ensuring that the mass of the predicted trajectory aligns with the mass of the ground truth at each time step. We have $L_{mass}=|\phi(x(t))-\phi(x'(t))|^2$, where $\phi$ is the summation of all chemical components.

\begin{figure}[t]
	\centering
		\centerline{\includegraphics[width=\textwidth]{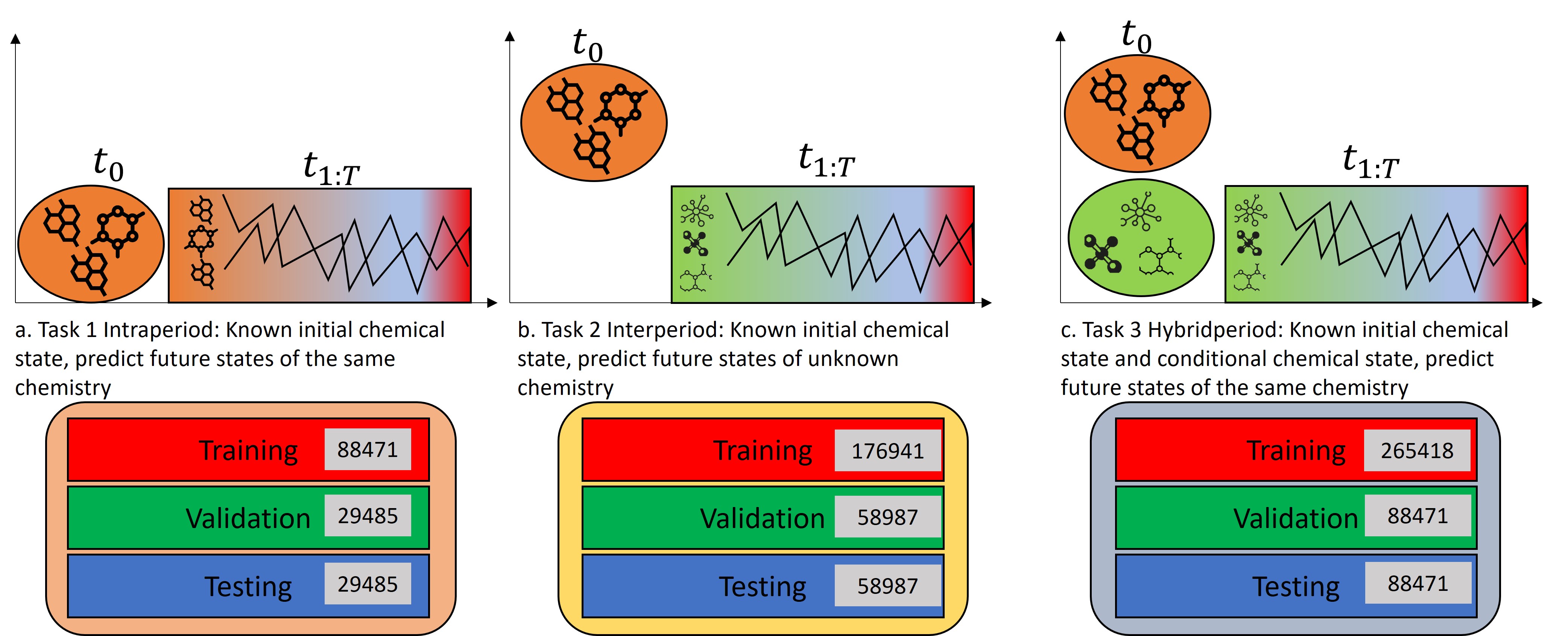}}
		\caption{\small{\textbf{Three chemical prediction task for model evaluation.} 
		}}
		\label{fig:task}
\end{figure}

\begin{table}[t]
\centering
\resizebox{\linewidth}{!}{
\begin{tabular}{clllccc}
\toprule
Tasks & Training & Validation & Testing & Environment input & Chemical input & Chemical output \\ \midrule
Task 1 & 88471 & 29485 & 29485 & 3 & 48 & 48 \\
Task 2 & 176941 & 58987 & 58987 & 3 & 48 & 100 \\
Task 3 & 265418 & 88471 & 88471 & 3 & 148 & 300 \\
\bottomrule
\end{tabular}
}
\caption{\textbf{Data summary of three tasks.}}
\label{tab:data}
\end{table}

\subsubsection{Evaluation}
To compare the estimations of the proposed \net against others, we apply the mean absolute errors (MAE) and the root mean squared errors (RMSE) to express the average difference. We also use the mean bias error (MBE) to calculate the estimation bias. The analytic equations for the estimation are:

\begin{equation}
\begin{matrix}
    \!\begin{aligned}
	& MAE = \sum_{i=1}^N \frac{|X_i-Y_i|}{N} \\
	& RMSE = \sqrt{\sum_{i=1}^N \frac{(X_i-Y_i)^2}{N}} \\
        & MBE = \sum_{i=1}^N \frac{X_i-Y_i}{N}
	\label{eq:eval}
	\end{aligned}
\end{matrix} 
\end{equation}

Furthermore, we also calculate the running time of the proposed model to compare with numerical simulation to see its efficiency, which could indicate how it can be utilized for large-scale modelling.

\section{Experiments}\label{experiments}
\subsection{Implementation Details}
\noindent $\bullet$ \textbf{Tasks.} We consider that our proposed \net can be used as a universal solver that can learn the intra- and inter-correlations among different chemical compounds over time. To test its efficiency, we design three tasks to validate its performance in Figure~\ref{fig:task}.

\begin{itemize}
\item \textbf{Task 1: Intraperiod chemical prediction.} Given the same 49 chemical compounds, we take their initial values ($x_0^i, i=1,2,...,49$) and environmental parameters $k$ to predict their future states. We use the ARCA box to simulate one-hour changes in chemical concentration and sample the observations every 5 minutes. The objective of the \net is to predict the results of all 12 time steps, as $x^i_t, i=1,2,...,49, t=1,2,...,12$.
\item \textbf{Task 2: Interperiod chemical prediction.} Based on the simulated concentrations of ARCA (concentrations above a defined threshold), we select 100 significant chemical compounds that have high impacts on air quality. We train the proposed model to take the same 49 chemical values ($x^i_0, i=1,2,...,49$) and environmental parameters $k$, and predict the time evolutions of the selected 100 chemical compounds ($y^i_t, i=1,2,...,100, t=1,2,...,12$).
\item \textbf{Task 3: Hybridperiod chemical prediction.} Similarly, as Task 2, we further select 300 more new chemical compounds that are generated over time. This is a much more challenging task, and we need additional chemical information to help the model understand the chains of chemical reactions. We design the model to take 49 initial chemical values and 100 new chemical values from Task 2, to output another 300 new chemical compounds at different time steps as ($z^i_t, i=1,2,...,300, t=1,2,...,12$).
\end{itemize}

\noindent $\bullet$ \textbf{Datasets.} We collected data based on the description in Section 2.1. We ran the simulations on CSC computers\footnote{https://csc.fi/} for 12 hours. We randomly split the data into non-overlapped training, validation and testing datasets. The data size of all three tasks is summarized in Table~\ref{tab:data}. 

To standardize the data for neural network training, we first take the logarithm of true chemical output to the base 10, then we normalize all data by dividing the maximum value, approximately 31.5. For the environmental parameters, we take the maximum and minimum values to normalize them to [-1, 1]. To further increase the data variety, we apply data augmentation to the training set. Specifically, we randomly roll the time evolution of the observation as $\hat{y}_i^t = y_i^{t+\tau}, i=1,2,...,100, t=1,2,...,12$, where 

\begin{small}
\begin{equation}
\hat{y}_i^t=\left\{ \begin{array}{rcl} y_i^{t+\tau}~~~~~ & \mbox{for} & \tau<12-t \\ y_i^{t+\tau-12} & \mbox{for} & \tau>12-t \end{array}\right. \text{where }  i=1,2,..., t=1,2,...,12 
\label{eq:roll}
\end{equation}
\end{small}

\noindent $\bullet$ \textbf{Parameter setting.} We train \net using Adam optimizer with the learning rate of $1\times10^{-3}$ and halved around 20K iterations. The batch size is set to 4096 and \net is trained for 100k iterations (about 2 hours) on a PC with one NVIDIA V100 GPU using PyTorch deep learning platform. The weighting factors in the total loss are defined empirically as: $\alpha_1=1, \alpha_2=10, \alpha_3=10, \alpha_4=1, \alpha_5=0.001$.

\begin{figure}[t]
	\centering
		\centerline{\includegraphics[width=\textwidth]{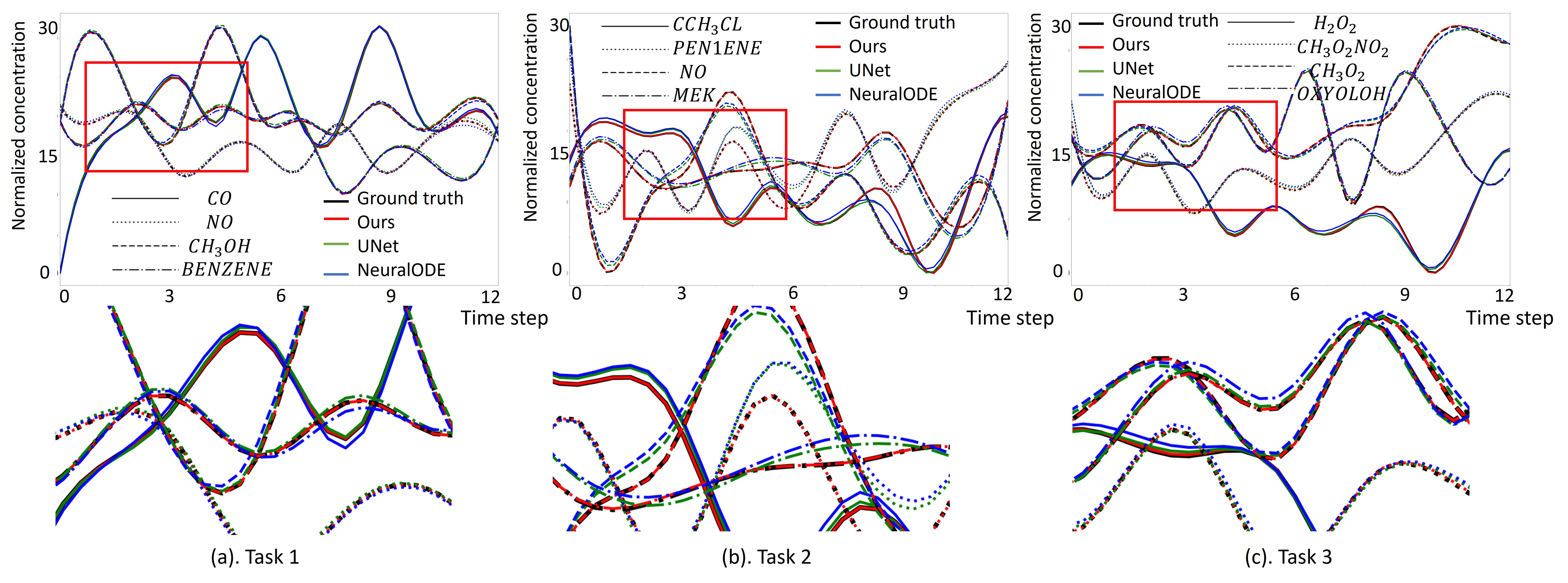}}
		\caption{\small{\textbf{Visualization of the time evolution of chemistry in Task 1, 2 and 3.} In (a), (b), and (c), we show the ground truth as red lines and the predictions with different colors. We pick four different chemical compounds for comparison, and we also enlarge the region in red boxes to highlight the prediction errors.
		}}
		\label{fig:vis_compare}
\end{figure}

\begin{table}[h]
\centering
\renewcommand\arraystretch{1.3}
\resizebox{\linewidth}{!}{
\begin{tabular}{clcccccccc}
\toprule
\multirow{2}{*}{Task} & \multirow{2}{*}{Model} & \multicolumn{3}{c}{Validaton} & \multicolumn{3}{c}{Testing} & Running time (s) & MACs (M)\\
 &  & RMSE & MAE & MBE & RMSE & MAE & MBE &  \\ \hline
\multirow{3}{*}{Task 1} & UNet & 0.4210 & 0.3292 & 0.2232 & 0.4267 & 0.3316 & 0.2152 & 5.15e-5 & 10.45 \\
 & NeuralODE & 0.3959 & 0.2741 & 0.0003 & 0.3958 & 0.2738 & 0.0002 & 1.06e-4 & 10.95 \\
 & \cellcolor{mistyrose}{Ours} & \cellcolor{mistyrose}{0.0194} & \cellcolor{mistyrose}{0.0086} & \cellcolor{mistyrose}{-1.9e-5} & \cellcolor{mistyrose}{0.0652} & \cellcolor{mistyrose}{0.0441} & \cellcolor{mistyrose}{-0.0028} & \cellcolor{mistyrose}{5.12e-5} & \cellcolor{mistyrose}{13.77}\\ \midrule
\multirow{3}{*}{Task 2} & UNet & 0.4283 & 0.2984 & 0.0163 & 0.4291 & 0.2988 & 0.0166 & 5.06e-5 & 14.11\\
 & NeuralODE & 0.2136 & 0.1244 & 0.0094 & 0.2144 & 0.1247 & 0.0074 & 1.26e-4 &12.13\\
 & \cellcolor{mistyrose}{Ours} & \cellcolor{mistyrose}{0.1156} & \cellcolor{mistyrose}{0.0312} & \cellcolor{mistyrose}{-0.0037} & \cellcolor{mistyrose}{0.1174} & \cellcolor{mistyrose}{0.0316} & \cellcolor{mistyrose}{-0.0037} & \cellcolor{mistyrose}{5.55e-5} & \cellcolor{mistyrose}{13.94}\\ \midrule
\multirow{3}{*}{Task 3} & UNet & 0.3096 & 0.1933 & -0.0009 & 0.3090 & 0.1932 & -0.0009 & 9.74e-5 & 45.03\\
 & NeuralODE & 0.2102 & 0.1079 & -0.0015 & 0.2086 & 0.1078 & -0.0016 & 2.18e-4 & 42.03\\
 & \cellcolor{mistyrose}{Ours} & \cellcolor{mistyrose}{0.0748} & \cellcolor{mistyrose}{0.0357} & \cellcolor{mistyrose}{0.0003} & \cellcolor{mistyrose}{0.0749} & \cellcolor{mistyrose}{0.0359} & \cellcolor{mistyrose}{0.0003} & \cellcolor{mistyrose}{1.07e-4} & \cellcolor{mistyrose}{44.34}\\ \bottomrule 
\end{tabular}
}
\caption{\textbf{Compare state-of-the-art methods on three tasks for chemical concentration prediction.} We test different methods on both validation and testing sets. For all metrics, the lower the values, the better prediction is achieved.}
\label{tab:sota}
\end{table}

\begin{figure}[b]
	\centering
		\centerline{\includegraphics[width=\textwidth]{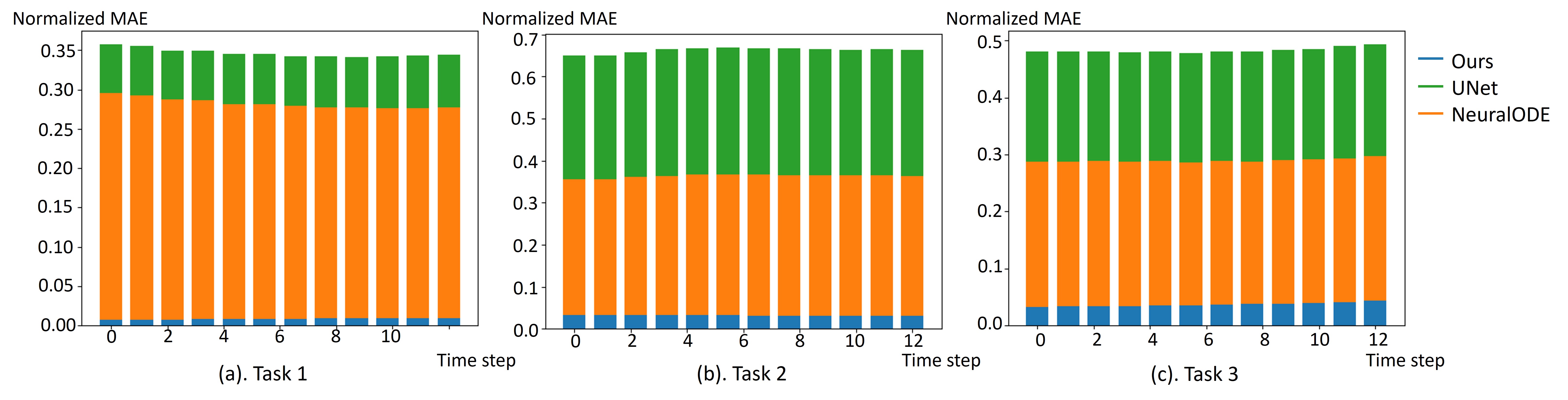}}
		\caption{\small{\textbf{Visualization of the mean errors of the time evolution in Task 1, 2 and 3.} We average all chemical compounds and show the mean absolute errors between ground truth and other predictions across different time steps.
		}}
		\label{fig:vis_time_compare}
\end{figure}

\subsection{Overall comparison with state-of-the-arts}
To demonstrate the efficiency of our proposed \net, we compare it with two state-of-the-art methods: UNet (\citet{unet}), NeuralODE (\citet{neuralode}), which are widely used in neural ODE/PDE processes. The comparison is shown in Table~\ref{tab:sota}. We compare different approaches on both validation and testing datasets in logarithm scales. We also show the model complexity by running time (seconds) and number of Multiply–accumulate operations (MACs). Compared to NeuralODE and UNet, we can observe that using our proposed \net can achieve the best performance in terms of RMSE, MAE and MBE. For instance, ours can improve the RMSE by about 0.28 $\sim$ 0.41 in Task 1, 0.1$\sim$ 0.3 in Task 2, and 0.14$\sim$ 0.23 in Task 3.  Comparing different tasks, we can see that the improvements from Task 1 to Task 3 are reduced which also indicates that the modelling difficulty is increased when more unknown chemical components are estimated. From the computation complexity, ours has a comparable number of operations as the other two and achieves a similar running speed. As the baseline, the numerical chemical model takes 0.05 seconds to run one simulation. It indicates that ours can be used as a fast emulator to accelerate the chemical modelling process. In practical applications, neural surrogate models can detect and respond to subtle climate changes far more quickly than traditional numerical models. This rapid detection has the potential to significantly reduce human and economic costs associated with catastrophic climate events. For visualization, we take different results on Tasks 1, 2, and 3 and visualize them in Figure~\ref{fig:vis_compare}. In tasks 1, 2, and 3, there are over 100 different chemical compounds, which makes it difficult to visualize them all in one figure. We randomly pick 4 individual chemical compounds and draw them in different line styles (solid, dotted, dashed, dash-dot). For each compound, we show the predictions of UNet, NeuralODE and ours in green, blue and red color. The ground truth value is shown in black color. We enlarge the red-box regions to demonstrate better the differences between ground truth and predictions. We can see that using ours can accurately align with the ground truth, while UNet and NeuralODE still produce distinct gaps between the ground truth. For a complete chemical compound comparison, you can find the figures in the supplementary material.

Meanwhile, we calculate the mean absolute prediction errors across different chemical compounds and summarize errors by time steps. In Figure~\ref{fig:vis_time_compare}, we compare different approaches to three different tasks. We can see that ours achieves the lowest errors across different time steps in all tasks. We can also observe that the further time steps, the higher the errors we get, which fits our assumption that long-term chemical prediction produces higher uncertainty. Depending on the difficulty of different tasks, we can also see that task 2 has the overall highest errors, and task 1 has the lowest errors. It is interesting to note that Task 3 uses 100 chemical compounds actively involved in the chemical reactions as additional input. This allows the model to better understand the correlations between the initial chemical values and the 300 new chemical compounds. For instance, knowing the concentration of \textit{OH} (one of the 100 significant chemical compounds for air quality) can help predict the concentration changes of \textit{C7300H} and \textit{OXYLOOH} (two of the 300 predicted chemical compounds). As a result, the model for Task 3 performs slightly better than that for Task 2.

\begin{table}[t]
\centering
\renewcommand\arraystretch{1.3}
\resizebox{\linewidth}{!}{
\begin{tabular}{cccccccccccccc}
\toprule
\multicolumn{5}{c}{Components} & \multicolumn{3}{c}{Task 1} & \multicolumn{3}{c}{Task 2} & \multicolumn{3}{c}{Task 3} \\ \midrule
AE & Attn & Time emb & Sinusoid & FNO & RMSE & MAE & MBE & RMSE & MAE & MBE & RMSE & MAE & MBE \\ \midrule
\checkmark &  &  &  &  & 0.3776 & 0.1876 & 0.0155 & 0.3976 & 0.2995 & 0.0368 & 0.5113 & 0.4010 & 0.0301 \\
\checkmark & \checkmark &  &  &  & 0.1252 & 0.0662 & 0.0044 & 0.2615 & 0.1589 & 0.0152 & 0.2455 & 0.1675 & 0.0089 \\
\checkmark &  & \checkmark &  &  & 0.3240 & 0.1665 & 0.0147 & 0.3589 & 0.2745 & 0.0235 & 0.5012 & 0.3370 & 0.0125 \\
\checkmark &  &  & \checkmark &  & 0.2256 & 0.1825 & 0.0035 & 0.1956 & 0.0899 & 0.0168 & 0.4888 & 0.3412 & 0.0115 \\
\checkmark &  &  &  & \checkmark & 0.0707 & 0.0486 & 0.0033 & 0.1389 & 0.0785 & -0.0082 & 0.0789 & 0.0386 & -0.0023 \\
\checkmark & \checkmark & \checkmark &  &  & 0.0649 & 0.0441 & -0.0029 & 0.1373 & 0.0675 & 0.0068 & 0.0768 & 0.0375 & -0.0010 \\
\checkmark & \checkmark & \checkmark & \checkmark &  & 0.0201 & 0.0109 & 0.0041 & 0.1182 & 0.0324 & 0.0031 & 0.1201 & 0.0328 & 0.0032 \\
\cellcolor{mistyrose}{\checkmark} & \cellcolor{mistyrose}{\checkmark} & \cellcolor{mistyrose}{\checkmark} & \cellcolor{mistyrose}{\checkmark} & \cellcolor{mistyrose}{\checkmark} & \cellcolor{mistyrose}{0.0194} & \cellcolor{mistyrose}{0.0086} & \cellcolor{mistyrose}{-1.9e-5} & \cellcolor{mistyrose}{0.1156} & \cellcolor{mistyrose}{0.0312} & \cellcolor{mistyrose}{-0.0037} & \cellcolor{mistyrose}{0.0748} & \cellcolor{mistyrose}{0.0357} & \cellcolor{mistyrose}{0.0003} \\ \bottomrule
\end{tabular}%
}
\caption{\textbf{Comparison on different key components of our proposed \net.} We report the results on validation datasets and for all metrics, lower values mean better performance.}
\label{tab:ablation}
\end{table}

\begin{table}[t]
\centering
\renewcommand\arraystretch{1.3}
\resizebox{\linewidth}{!}{
\begin{tabular}{clccccc}
\toprule
\multicolumn{2}{c}{Task} & MSE (baseline) & MSE+Derivs & MSE+Derivs+Idn & MSE+Derivs+Mass & \cellcolor{mistyrose}{MSE+Derivs+Idn+Mass} \\ \midrule
\multicolumn{1}{c}{\multirow{3}{*}{Task 1}} & RMSE & 0.0315 & 0.0268 & 0.0225 & 0.0210 & \cellcolor{mistyrose}{0.0194} \\
\multicolumn{1}{c}{} & MAE & 0.0126 & 0.0102 & 0.0094 & 0.0089 & \cellcolor{mistyrose}{0.0086} \\
\multicolumn{1}{c}{} & MBE & 0.0008 & 2.9e-5 & 3.4e-5 & 2.8e-5 & \cellcolor{mistyrose}{-1.9e-5} \\ \midrule
\multicolumn{1}{c}{\multirow{3}{*}{Task 2}} & RMSE & 0.1876 & 0.1820 & 0.1502 & 0.1501 & \cellcolor{mistyrose}{0.1156} \\
\multicolumn{1}{c}{} & MAE & 0.0589 & 0.0566 & 0.0408 & 0.0386 & \cellcolor{mistyrose}{0.0312} \\
\multicolumn{1}{c}{} & MBE & 0.0089 & 0.0091 & -0.0076 & 0.0064 & \cellcolor{mistyrose}{-0.0037} \\ \midrule
\multicolumn{1}{c}{\multirow{3}{*}{Task 3}} & RMSE & 0.0896 & 0.0895 & 0.0820 & 0.0766 & \cellcolor{mistyrose}{0.0748} \\
\multicolumn{1}{c}{} & MAE & 0.0523 & 0.0510 & 0.0461 & 0.0447 & \cellcolor{mistyrose}{0.0357} \\
\multicolumn{1}{c}{} & MBE & 0.0015 & 0.0015 & 0.0008 & 0.0010 & \cellcolor{mistyrose}{0.0003} \\ \bottomrule 
\end{tabular}%
}
\caption{\textbf{Comparison on different loss terms of our proposed \net.} We report the results on the validation dataset and we can see that the combination of all losses achieves the best performance.}
\label{tab:loss_ablation}
\end{table}

\subsection{Ablation studies}
We conduct several ablations to test the key components of the proposed \net and report the results in Table~\ref{tab:ablation}.

\begin{figure}[t]
	\centering
		\centerline{\includegraphics[width=\textwidth]{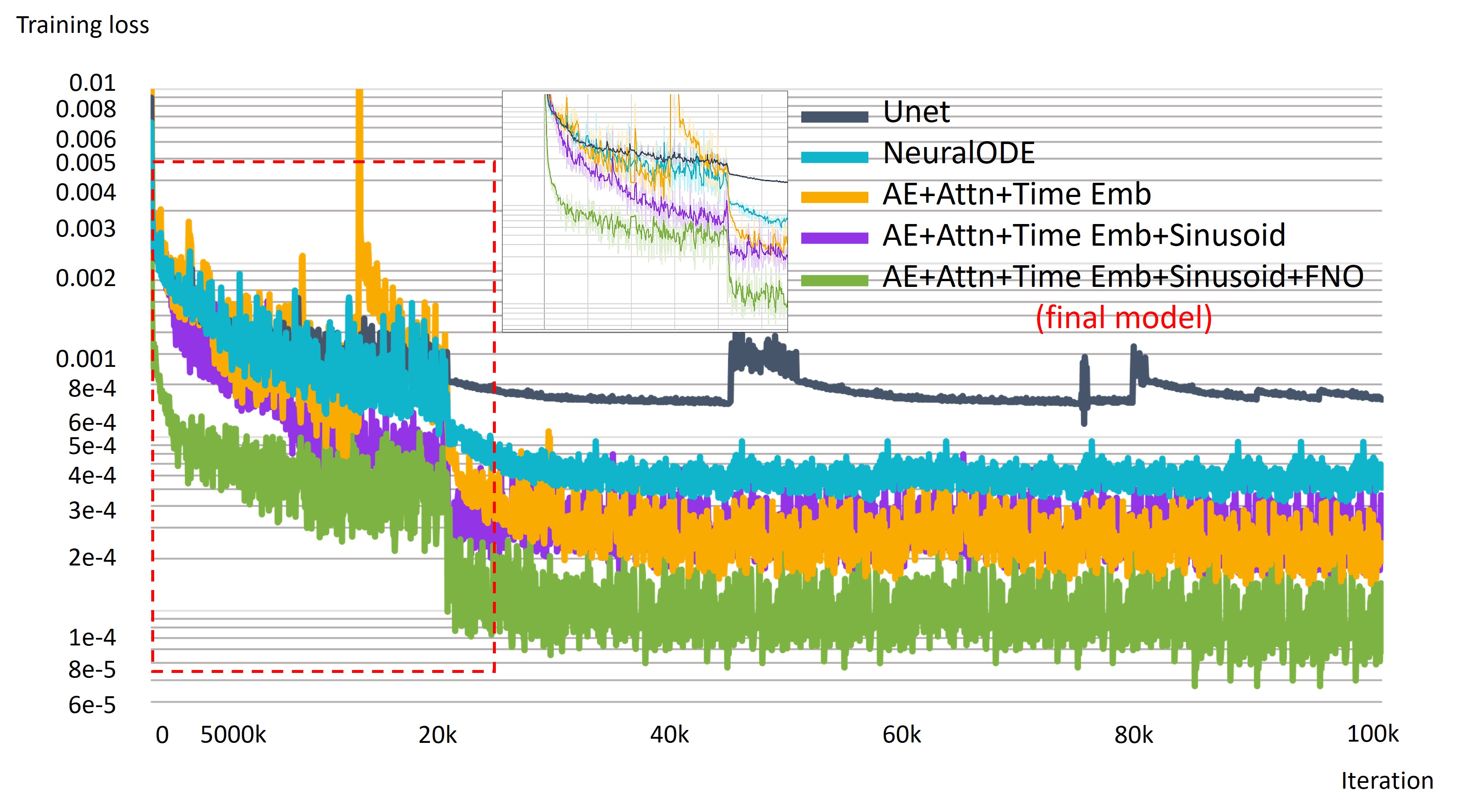}}
		\caption{\small{\textbf{Training loss comparison among ours and others.} We show different approaches in different colors. We enlarge the red-boxed region and display it at the upper center. We can see the improvements of using our proposed \net in both convergence speed and final loss. 
		}}
		\label{fig:train_loss}
\end{figure}

We use the validation dataset of all three Tasks to demonstrate the effects of different key components, including AE (AutoEncoder), attention (attn), time embedding (Time emb), Sinusoid, FNO, and their combinations. Using AE is our baseline. We can see that the last row is our final model, which shows the best performances in all metrics. Individually, we can see that using attention and FNO can achieve the most improvements by approximately 0.1 $\sim$ 0.4 in terms of RMSE. Using time embedding and Sinusoid can also improve the RMSE by about 0.01 $\sim$ 0.2. 

To further demonstrate the effect of the key components of attention and FNO, we visualize the training loss convergences in Figure~\ref{fig:train_loss}. We enlarge the region within the red box and highlight it at the upper center to illustrate the loss differences. Compared to Unet and NeuralODE, we can see that ours with attention and/or FNO achieves the fastest convergence speed and lowest training loss. 

\begin{figure}[t]
	\centering
		\centerline{\includegraphics[width=\textwidth]{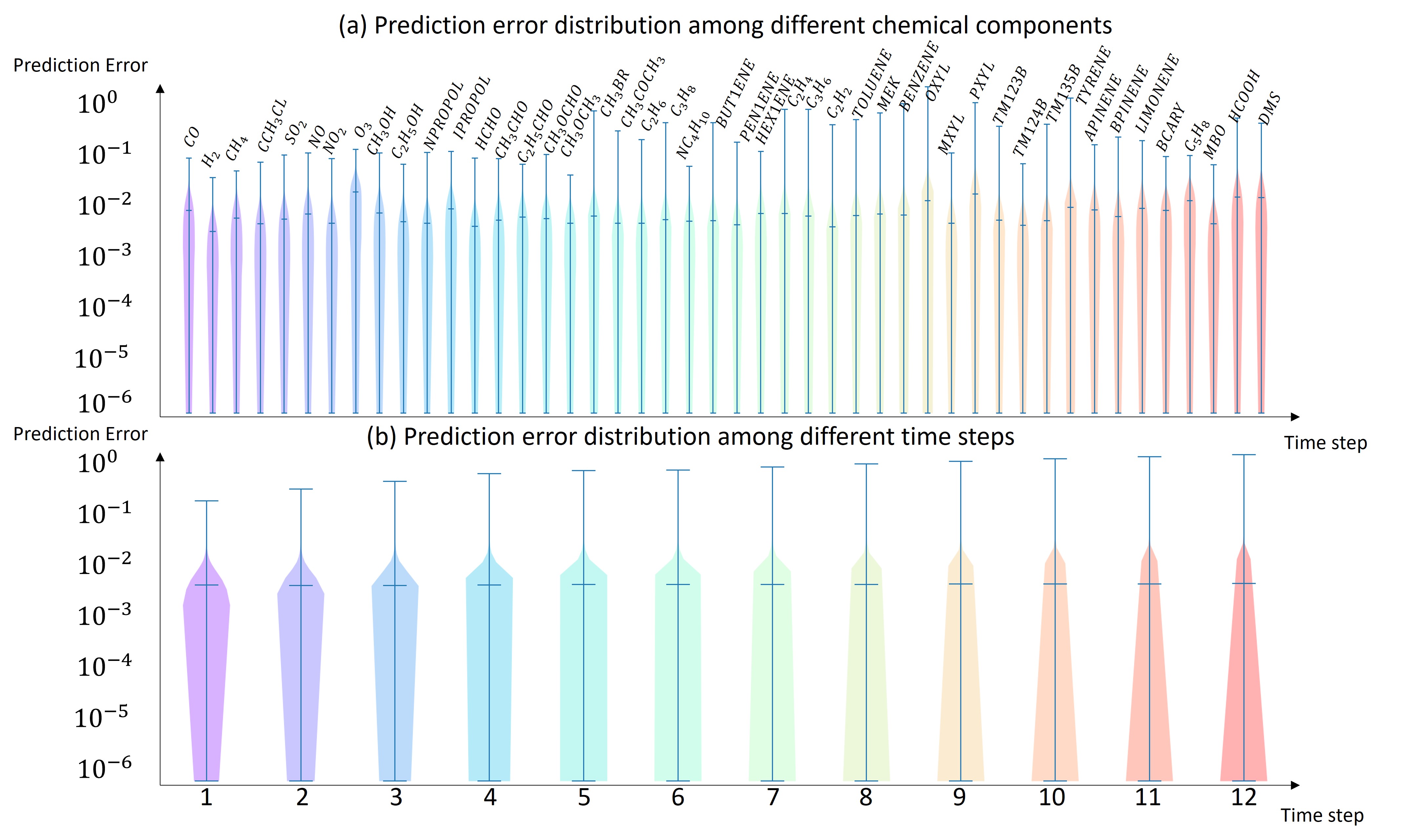}}
		\caption{\small{\textbf{Validation error distribution visualization via violin plots.} We show the error distributions (using the logarithm scale for better visualization) of different chemical components (a), and different time steps (b). The narrower the color shades, the smaller the error distributions.
		}}
		\label{fig:error_task1}
\end{figure}

\begin{figure}[t]
	\centering
		\centerline{\includegraphics[width=\textwidth]{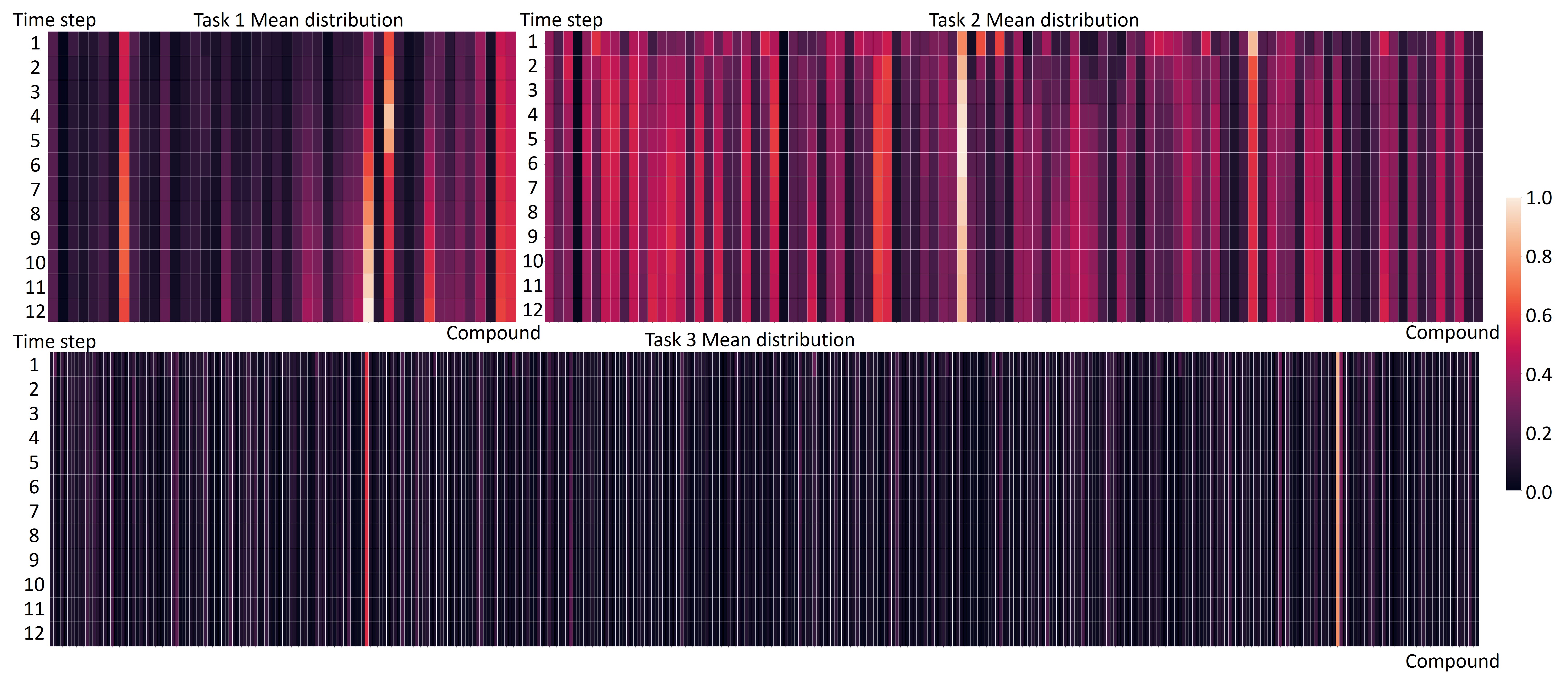}}
		\caption{\small{\textbf{Mean distribution of validation errors of three tasks.} We show the mean values of error distributions on three tasks, where the horizontal axis represents different chemical components and the vertical axis represents different time steps.
		}}
		\label{fig:mean_error}
\end{figure}

\begin{figure}[t]
	\centering
		\centerline{\includegraphics[width=\textwidth]{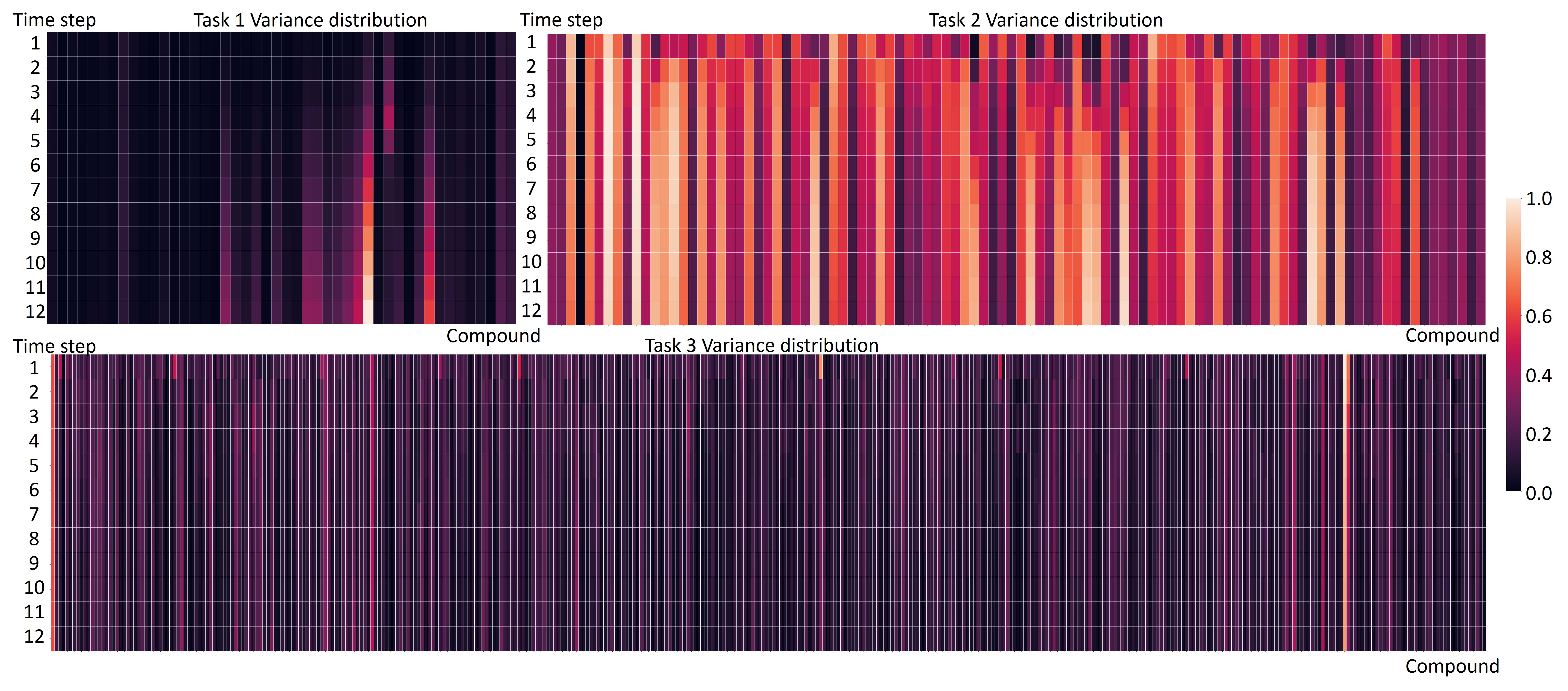}}
		\caption{\small{\textbf{Variance distribution of validation errors of three tasks.} We show the variance values of error distributions on three tasks, where the horizontal axis represents different chemical components and the vertical axis represents different time steps.
		}}
		\label{fig:var_error}
\end{figure}

Finally, we train our proposed \net with different loss combinations, so that we can validate the effects of using physics-informed losses for model optimization. As the baseline, we train the model with MSE loss ($L_{recon}$), and then we individually add other losses to retrain the model, including first- and second-order derivative losses (Derivs), identity loss (IDN), and mass conservation loss (Mass). In Table~\ref{tab:loss_ablation}, we show the RMSE, MAE, and MBE results on three tasks. Our observations are: 1) from columns 2 and 3, we can see that using identity and mass conservation losses have a more significant loss drop, about 0.03 $\sim$ 0.1 in RMSE, 2) using derivative loss has more visible effects on Task 1 than Task 2 and 3 because Task 1 is intraperiod chemical prediction, that is, the derivative can better constrain the chemical reaction process for accurate estimation. 

\subsection{Statistical analysis}
To study in depth the proposed \net on the ability of chemical modelling, we study the error distributions among different chemical components and time steps, so that we can see whether the model can estimate the patterns for each individual component, as well as long-term regression. In Figure~\ref{fig:error_task1}, we show the error distribution in Task 1. We can see that 1) the model performs unevenly on different chemical components in terms of means and variances, and 2) the errors increase when predicting further steps in the future. 

For Tasks 2 and 3, we have a much larger number of chemical components, 100 to 300 chemical components, for computation. To efficiently demonstrate the model performance, we show the 2D heatmap of mean and variance distributions on the validation datasets. Figure~\ref{fig:mean_error} and Figure~\ref{fig:var_error} show the mean and variance of error distributions in all three tasks. The brighter the color, the higher the errors the model produces. From Figure~\ref{fig:mean_error}, we can observe that the average performance differences depend on the specific chemical components. From Figure~\ref{fig:var_error}, we can see that the error variances vary because of the time steps. The farther the future step to predict, the higher the variances the model gets. The patterns also match our observations in Figure~\ref{fig:error_task1}.

Specifically, we are interested in the chemistry that the model fails to predict well. In Task 1, the top five highest errors come from $O_3, OXYL, PXYL$, $HCOOH, DMS$.  Ozone ($O_3$) is one of the main oxidants in the atmosphere and reacts with many compounds. Therefore its predictions are related to the concentrations of many other compounds. 1,2-dimetylbenzene ($OXYL$) and 1,4-dimethylbenzene ($PXYL$) are aromatics which react mainly with the hydroxyl radical ($OH$). OH is formed inside the model and is the most important oxidant of the atmosphere during daytime and reactive towards nearly all compounds. The uncertainty for $OXYL$ and $PXYL$ is mainly related to the performance of OH and the same is the reason for dimethylsulphide ($DMS$) and formic acid ($HCOOH$). For Task 2, the top four highest errors come from $HOC_2H_4CO_2H, PXYLCO_2H, KLIMONIC, C_{88}CO_2H$. For Task 3, the top two highest errors come from $OXYQOOH, LMLKBCO$. All these compounds are higher-order reaction products of our initial compounds and are formed after several reactions with ozone, the nitrate and the hydroxyl radical. The most promising way to decrease the uncertainties of these compounds is to increase our training data set in the future.

\section{Conclusion}\label{sec5}
In this paper, we propose the first work on a neural network emulator for fast chemical modelling, dubbed \net. We utilize the numerical simulation to generate a large-scale chemical dataset which is used to train the proposed \net, so that it can learn the hidden inter- and intra-correlations among chemical molecules. We propose to combine attention and neural ODE operations to model the time-dependent chemical reactions. Meanwhile, we lift the neural ODE as a Fourier domain convolution such that we can efficiently model the global continuous ODE. The time embedding and Sinusoid operators also help model the oscillation patterns to mimic the concentration changes of chemistry. To demonstrate the efficiency and effectiveness of the proposed model, we test it on three tasks, including intraperiod, interperiod and hybridperiod chemical predictions. Extensive experiments show that ours achieves the best performance in both accuracy and running speed. This work paves a new direction in AI for atmospheric chemistry, and we will continue to explore graph neural networks, equivalent neural networks, and other advanced models to integrate chemical knowledge for physics-informed processing. 

\bibliographystyle{elsarticle-harv} 
\bibliography{main}

\end{document}


\begin{frontmatter}



\title{Supplementary material: Neural Network Emulator for Atmospheric Chemical ODE} 

\author[a,b]{Zhi-Song Liu}
\author[c]{Petri Clusius}
\author[b,c]{Michael Boy}

\affiliation[a]{organization={School of Engineering Science, Lappeenranta-Lahti University of Technology LUT},
            city={Lahti},
            postcode={15110},
            country={Finland}}
            
\affiliation[b]{organization={Atmospheric Modelling Centre Lahti, Lahti University Campus},
            city={Lahti},
            postcode={15140},
            country={Finland}}
            
\affiliation[c]{organization={Institute for Atmospheric and Earth System Research (INAR), The University of Helsinki},
            city={Helsinki},
            postcode={00014},
            country={Finland}}



\end{frontmatter}

\section{Network architecture}
The network detail is shown in Figure~\ref{fig:network_detail}. We show the dimension of feature maps to indicate the computation process. $N$ is the number of chemical compounds. $B$ is the batch size. For different tasks, we have different $N$. As introduced in Section 3.2, the encoder consists of several MLP layers followed by Sine activation functions. It takes as input the initial chemical concentration value and the environmental factors to learn a joint hidden feature vector $B\times 256$. Then we expand and repeat it to $B\times 256 \times 12$, where 12 is the total time steps it learns to predict. A self-attention module learns to extract non-local features for better feature representation. The sinusoid mapping projects the time code to the dimension of $B\times 256\times 12$, then we multiply it with the feature vector and pass it to the LinearODE module. The linearODE module is made of several FNO blocks, which process the input feature by global convolution. Finally, the decoder also contains a set of MLP layers with sine activation functions to produce final time prediction.

\begin{figure}[t]
	\centering
		\centerline{\includegraphics[width=\textwidth]{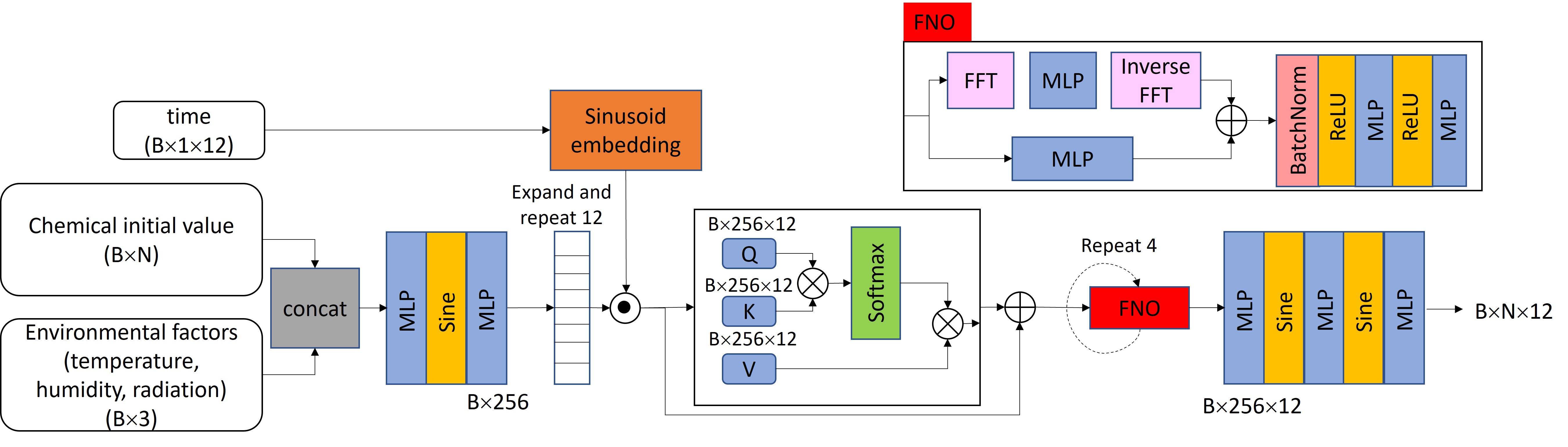}}
		\caption{\small{\textbf{The detail of the proposed \net for chemical concentration prediction.} We show the dimension of feature maps to indicate the computation process. The encoder contains a set of MLP layers with sine activation functions. The output hidden feature is further processed by the self attention module. The decoder also consists of several MLP layers with sine activations. In the middle, we use 4 FNO blocks to learn the inter- and intra-correlations among different chemical reactions and time evolutions.
		}}
		\label{fig:network_detail}
\end{figure}

\section{Visualization of chemical prediction}
In Section 4.2 of the main manuscript, we show the prediction comparisons of four individual chemical compounds. In Section 4.3, we also show statistical comparisons among different methods. However, it would be interesting to show an overall comparison of all chemical compounds. Please note that we have three different tasks, and each task processes different numbers of chemical compounds. For task 1, the proposed \net predicts 49 different chemical elements. For tasks 2 and 3, it needs to predict 100 and 300 chemical compounds. In the following figures, we use dashed black lines to represent the ground truth data, and solid color lines to represent our predictions. We can see that using ours can accurately align with the ground truth. For Task 2 and 3, we show the results in Figure 4. We can see that the color lines are overlapped with the dashed black lines, which indicates that our predictions align well with the ground truth values.

\begin{sidewaysfigure}
	\centering
		\centerline{\includegraphics[width=1.3\textwidth]{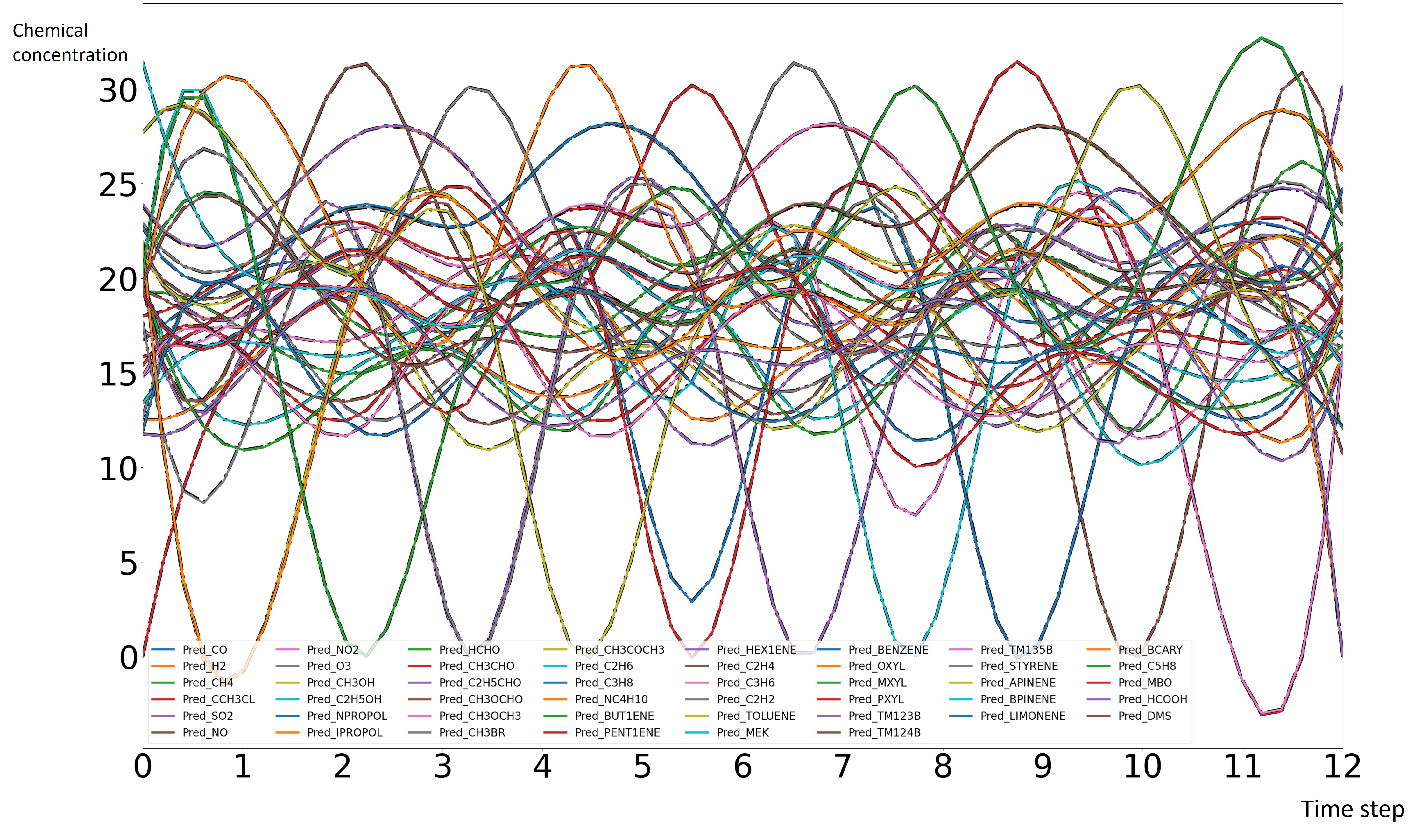}}
		\caption{\small{\textbf{The overall chemical concentration prediction on Task 1.} We show the ground truth as dashed black lines and the predictions as solid lines with different colors. 
		}}
		\label{fig:task1}
\end{sidewaysfigure}

\begin{sidewaysfigure}
	\centering
		\centerline{\includegraphics[width=1.3\textwidth]{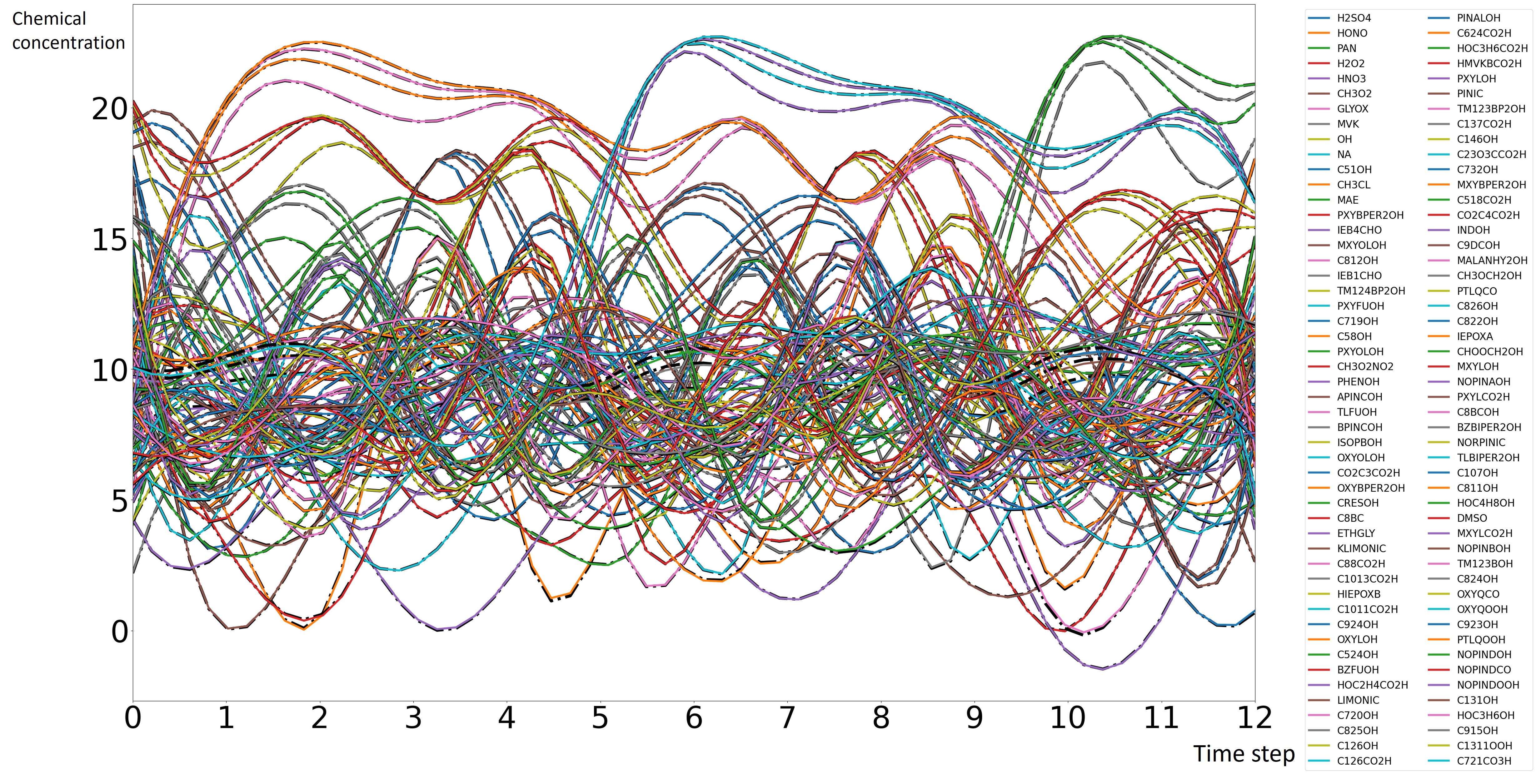}}
		\caption{\small{\textbf{The overall chemical concentration prediction on Task 2.} We show the ground truth as dashed black lines and the predictions as solid lines with different colors. 
		}}
		\label{fig:task2}
\end{sidewaysfigure}

\begin{sidewaysfigure}
	\centering
		\centerline{\includegraphics[width=1.3\textwidth]{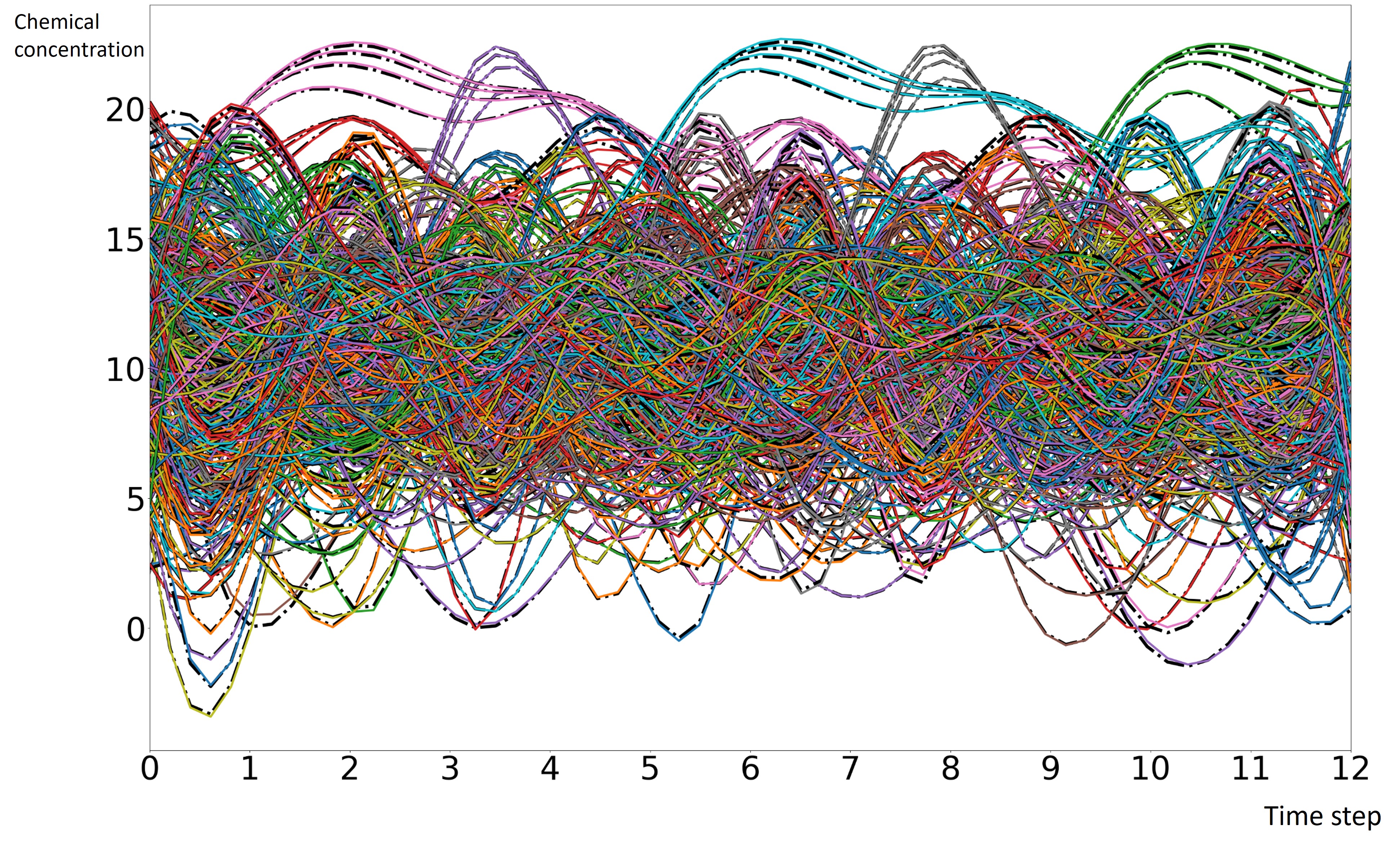}}
		\caption{\small{\textbf{The overall chemical concentration prediction on Task 3 [Data].} We show the ground truth as dashed black lines and the predictions as solid lines with different colors. 
		}}
		\label{fig:task3}
\end{sidewaysfigure}

\begin{figure}[t]
	\centering
		\centerline{\includegraphics[width=1.2\textwidth]{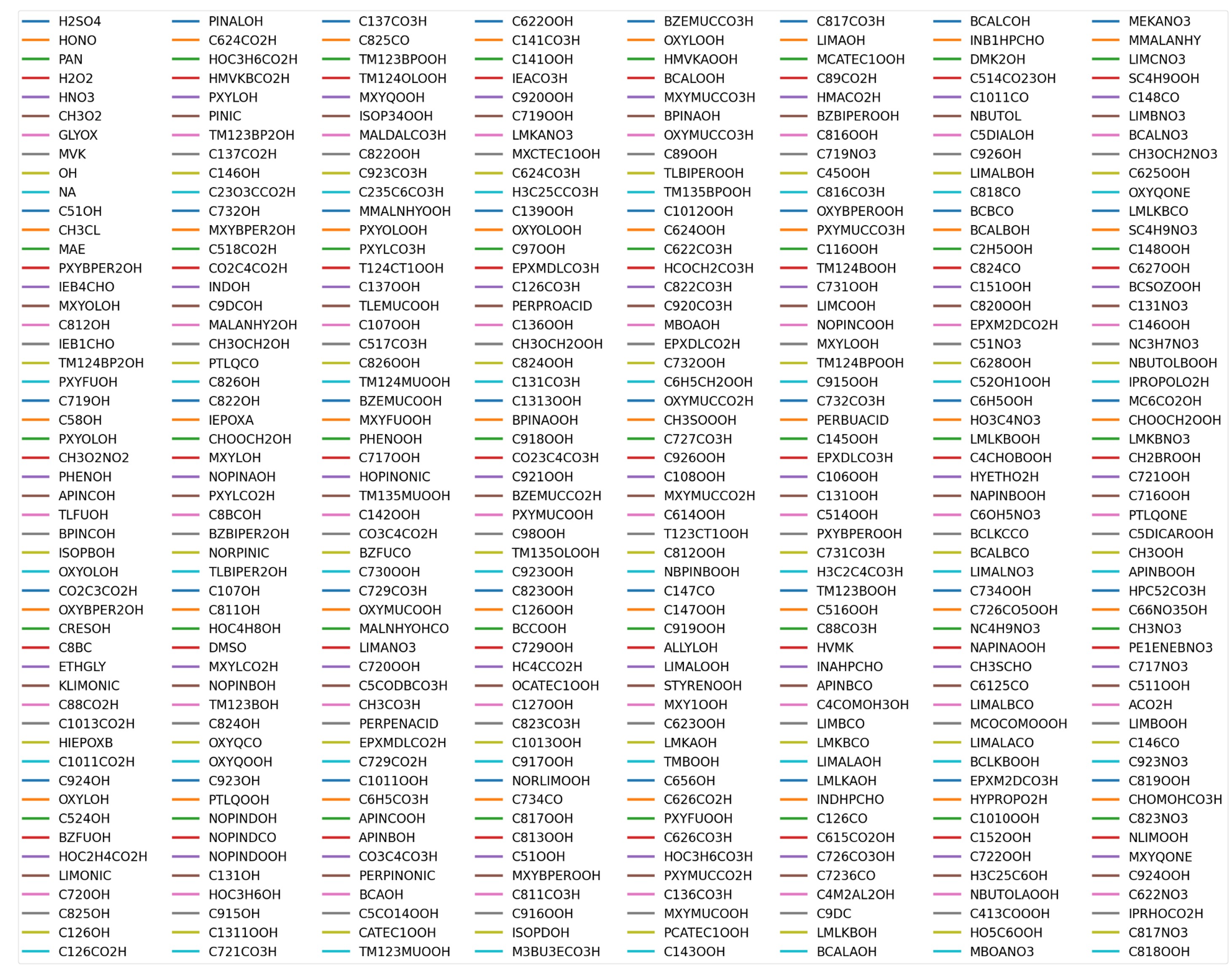}}
		\caption{\small{\textbf{The overall chemical concentration prediction on Task 3 [Label].} The label information for Task 3.
		}}
		\label{fig:network_detail}
\end{figure}